\documentclass[]{arxiv_preprint}
\usepackage{amsmath,amsfonts,amssymb,array}
\usepackage{xspace}
\usepackage{needspace}
\usepackage{tikz}
\usetikzlibrary{arrows.meta,positioning,calc,fit}

\definecolor{cvprblue}{rgb}{0.21,0.49,0.74}
\definecolor{pipelineblue}{HTML}{EAF1F8}
\definecolor{pipelinegreen}{HTML}{EAF4EE}
\definecolor{pipelineorange}{HTML}{FAEFE4}
\definecolor{draftcolor}{HTML}{8C4C17}
\newcommand{\egoloaf}{EgoLoAF}
\newcommand{\pifive}{\ensuremath{\pi_{0.5}}}

\newcommand{\pose}{\phi}
\newcommand{\trans}[2]{{}^{#1}\!T_{#2}}

\DeclareMathOperator{\med}{median}

\newcommand{\authornewline}[2][]{\addtolist[#1]{#2}{\authorlist}{\authorformat}{,\par}}

\title{\textcolor{metablue}{EgoLoAF}: Learning Humanoid Loco-Manipulation from Egocentric Demonstrations with Hand--Object Affordances}
\author[1,3,\ast,\dagger]{Yiming~Zeng}
\author[1,3,\ast]{Jiangyuxuan~Dai}
\author[1,2,3,\ast,\dagger]{Yuheng~Na}
\author[1,3]{TAN~ZHU~XUAN}
\author[1]{Zihan~Lin}
\author[1]{Xinfang~Qi}
\author[1]{Shijian~Zhang}
\authornewline[1]{Chengqi~Shi}
\author[1]{Yihan~Kuang}
\author[2]{Zhide~Zhong}
\author[2]{Junjie~He}
\author[2]{Junfeng~Li}
\author[1]{Zimu~Han}
\author[1,\ddagger]{Naipeng~Li}
\author[2,\ddagger]{Haoang~Li}
\affiliation[1]{Xi'an Jiaotong University}
\affiliation[2]{The Hong Kong University of Science and Technology (Guangzhou)}
\affiliation[3]{Xbotics Research Team}
\contribution[\ast]{Equal contribution}
\contribution[\dagger]{Project leads}
\contribution[\ddagger]{Corresponding authors}
\abstract{Training humanoid vision-language-action (VLA) models relies heavily on costly real-robot teleoperation.
Recent efforts convert egocentric human demonstrations into humanoid training data.
However, most methods still require wearable motion trackers and leave important hand--object interaction cues implicit.
Many also mix in real-robot teleoperation data for VLA post-training.
These dependencies limit scalable data collection, while implicit interaction cues make precise contact harder to learn.
To overcome these limitations, we propose \textbf{\egoloaf{}}, the first paradigm to derive all training data for humanoid loco-manipulation directly from egocentric demonstrations without wearable motion trackers.
We post-train the VLA on the converted data alone, without real-robot teleoperation demonstrations, and use explicit hand--object interaction cues to guide action generation.
Specifically, to obtain humanoid action targets, we estimate torso motion from recorded camera poses and reconstruct both hand trajectories from video.
From the same recording, we use each hand's current pose relative to the object to represent hand--object affordances.
During post-training, we inject these affordances into the action expert through cross-attention while keeping the vision-language backbone frozen.
Additionally, we release 128,400 egocentric demonstration episodes spanning 1,000 hours and 1,284 tasks, together with converted humanoid action trajectories and rendered robot-view observations.
Across four real-world tasks, \egoloaf{} nearly achieves teleoperation performance (73.1 vs.\ 74.4 in mean Task Progress Score), while achieving a mean $4.03\times$ improvement in data-collection efficiency.
}
\metadata[Project Page]{\url{https://egoloaf.github.io/}}
\hypersetup{pdftitle={EgoLoAF: Learning Humanoid Loco-Manipulation from Egocentric Demonstrations with Hand-Object Affordances},pdfauthor={Yiming Zeng, Jiangyuxuan Dai, Yuheng Na, TAN ZHU XUAN, Zihan Lin, Xinfang Qi, Shijian Zhang, Chengqi Shi, Yihan Kuang, Zhide Zhong, Junjie He, Junfeng Li, Zimu Han, Naipeng Li, Haoang Li}}

\begin{document}
\maketitle
\begin{figure}[htbp]
\centering

\includegraphics[width=\linewidth]{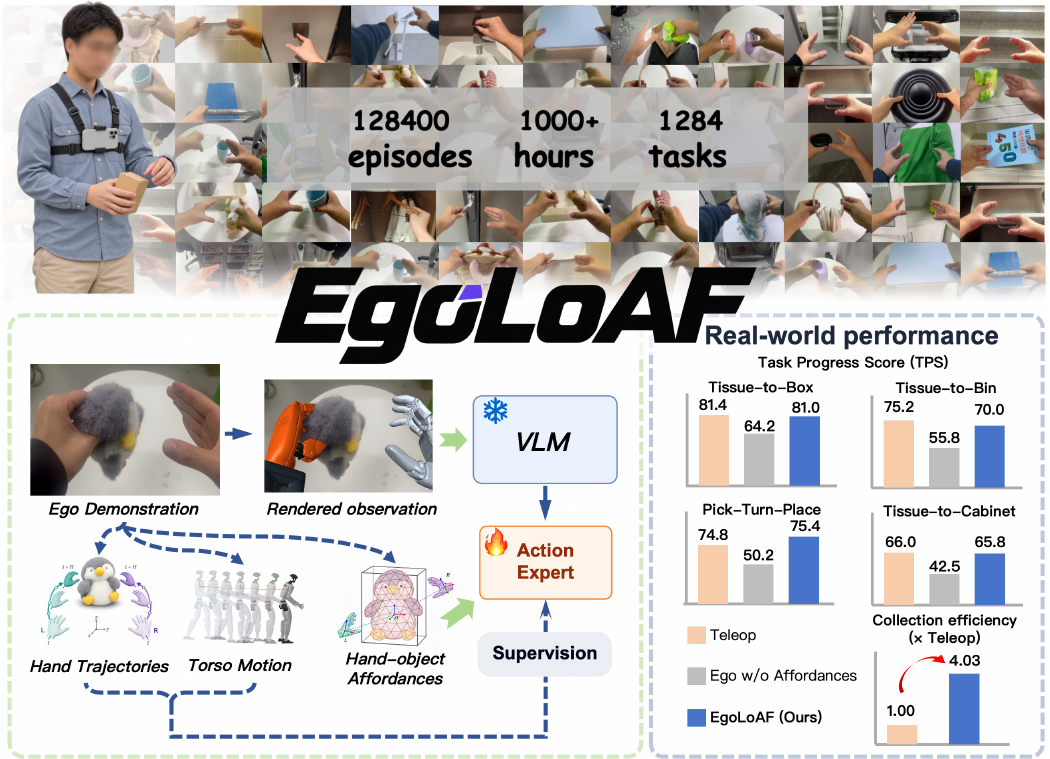}
\caption{\textbf{Overview of \egoloaf{}, a novel data collection paradigm for humanoids.}
\textbf{Top:} We release 128,400 egocentric demonstration episodes spanning 1,000 hours and 1,284 tasks, collected without wearable motion trackers.
\textbf{Bottom left:} We convert these demonstrations into rendered robot-view observations, torso and hand action targets, and current hand--object affordances. The action targets supervise the action expert, while the affordances condition action generation. Post-training uses no real-robot teleoperation demonstrations, and the vision-language model (VLM) remains frozen.
\textbf{Bottom right:} Across four real-world tasks, \egoloaf{} nearly matches the teleoperation-trained baseline in mean Task Progress Score (TPS; 73.1 vs.\ 74.4), while \textbf{more than quadrupling data-collection efficiency on average ($4.03\times$)}.}
\label{fig:teaser}
\end{figure}

\section{Introduction}
\label{sec:intro}

Vision-language-action (VLA) models and related policies have made progress in robot manipulation and humanoid loco-manipulation \citep{wholebodyvla,openhlm,lambda0,evomemvla,hilumi,sphlm,hipolicy}.
However, training humanoid VLAs still relies heavily on real-robot teleoperation.
Each demonstration requires a skilled operator to teleoperate a physical humanoid.
Resetting the robot and workspace between trials further increases collection time.
This makes it costly to collect demonstrations across many tasks and environments.
To address this challenge, egocentric human data offers an alternative.

Recently, there have been efforts to convert egocentric human demonstrations into training data for humanoid loco-manipulation \citep{egohumanoid,egohumanoidv2,lambda0,koda}.
However, existing approaches still face two limitations.
First, most of these methods use wearable motion trackers to obtain humanoid action labels.
Each recording session involves fitting the trackers to the demonstrator and synchronizing their motion measurements with the video.
These setup requirements make data collection more difficult to scale.
Second, most of these methods use converted human trajectories as action targets and leave the current hand--object relationship implicit.
This makes precise contact harder because the policy must infer where the robot's hands are relative to the object before deciding how to move.
If the robot misjudges the hand--object relationship, it will reach the right object but still miss the correct contact required by the task.

To overcome these limitations, we propose \egoloaf{}, a paradigm that converts egocentric human demonstrations into humanoid loco-manipulation training data without wearable motion trackers.
The same demonstrations also provide hand--object affordances that guide VLA post-training (Fig.~\ref{fig:teaser}).
Specifically, we reconstruct both hand trajectories from egocentric video and derive a torso trajectory from the recorded camera poses.
We use the current torso position and orientation as a reference to describe subsequent torso and hand motion as humanoid action targets.
We also estimate object poses from the same observations and represent hand--object affordances using each hand's current pose relative to the object.
To align training observations with the humanoid, we replace human arms and hands in the video with rendered robot counterparts.
Then, we use the rendered observations and action targets for humanoid VLA post-training, and inject hand--object affordances into the action expert through cross-attention.
During post-training, we update the action expert while keeping the VLM backbone frozen.

Using this egocentric-to-humanoid conversion pipeline, we have constructed and released the \egoloaf{} dataset, containing 128,400 egocentric demonstration episodes spanning 1,000 hours and 1,284 tasks.
The dataset includes converted humanoid action trajectories and rendered observations for direct use in humanoid VLA training.
In our experiments, \egoloaf{} policies are post-trained only on egocentric demonstrations without real-robot teleoperation data.
Across four real-world tasks, \egoloaf{} achieves a mean Task Progress Score (TPS) of 73.1, close to 74.4 for real-robot teleoperation.
At the same time, \egoloaf{} collects training data $4.03\times$ as efficiently as teleoperation on average.

\Needspace{6\baselineskip}
Our contributions are as follows.
\begin{enumerate}
    \item We propose \egoloaf{}, a paradigm that converts egocentric human demonstrations into humanoid loco-manipulation training data without wearable motion trackers. Each demonstration provides action targets and hand--object affordances, enabling VLA post-training without real-robot teleoperation data.
    \item We release a dataset of 128,400 egocentric demonstration episodes spanning 1,000 hours and 1,284 tasks. Each episode includes converted humanoid action trajectories, rendered observations, and hand--object affordances for humanoid VLA training.
    \item We propose a post-training method that encodes current object-relative hand poses as interaction affordances and injects these affordances into the action expert through cross-attention, while keeping the VLM backbone frozen. Across four real-world tasks, \egoloaf{} nearly matches teleoperation performance (73.1 vs.\ 74.4 mean TPS) while \textbf{more than quadrupling data-collection efficiency on average ($4.03\times$)}.
\end{enumerate}

\section{Related work}
\label{sec:related}

\paragraph{Humanoid vision--language--action models.}
Humanoid VLAs learn language-guided loco-manipulation through transferable representations and structured action generation.
WholeBodyVLA \citep{wholebodyvla} and $\Psi_{0}$ \citep{psi0} learn from human videos before adaptation with humanoid trajectories.
OpenHLM \citep{openhlm} studies whole-body teleoperation, VLA adaptation, and heterogeneous co-training.
Action designs include body-part tokenization, hierarchical generation, and sparse-pose interfaces \citep{holom,hafvla,sphlm}.
These interfaces differ in how much whole-body motion the policy predicts and how much the controller supplies from compact task-space goals.
HEX \citep{hex} conditions actions with dynamics experts, while RISE \citep{rise} learns behavioral experts and their transitions.
Interaction-aware policies add whole-body tactile observations \citep{univlat} or persistent 3D object states for action generation and geometric verification \citep{potvla}.
\egoloaf{} constructs humanoid action supervision directly from egocentric recordings.

\paragraph{Humanoid learning from egocentric demonstrations.}
Egocentric transfer requires aligned observations and executable humanoid motion targets.
HAT \citep{hat} and In-N-On \citep{innon} use shared representations and retargeting for human--robot manipulation co-training.
EgoVLA \citep{egovla} adapts human hand actions to simulated humanoids using robot demonstrations and inverse kinematics.
EgoHumanoid \citep{egohumanoid} aligns observations and actions for loco-manipulation co-training.
EgoHumanoid-V2 \citep{egohumanoidv2} and EgoAlign \citep{egoalign} use wearable tracking and humanoid motion refinement to enable human-only target-task training.
KODA \citep{koda} combines tracker-free motion reconstruction, contact-aware hand refinement, and robot-view rendering, but mixes real-robot demonstrations into policy training.
Other interfaces address active head--hand coordination \citep{halomi} and whole-body contact goals \citep{pantact}.
$\lambda_{0}$ \citep{lambda0} scales aligned whole-body human supervision through HumanVerse and humanoid mid-training.
Transfer methods also differ in where they use geometry.
Refining a demonstrated trajectory improves the supervision available during training.
Explicit affordance conditioning provides complementary information: the current hand--object relationship from which the robot must act during deployment.
Motion refinement and affordance conditioning therefore affect different parts of the learning pipeline.
KODA's evaluated recordings include phone depth \citep{koda}; our annotation uses RGB and recorded camera poses, with depth estimated from images.
\egoloaf{} combines tracker-free motion annotation with post-training on converted egocentric data alone.
Each recording pairs future torso and hand action targets with current object-relative hand poses as affordance conditions.

\section{Method}
\label{sec:method}
\begin{figure}[htbp]
\centering
\includegraphics[width=\textwidth]{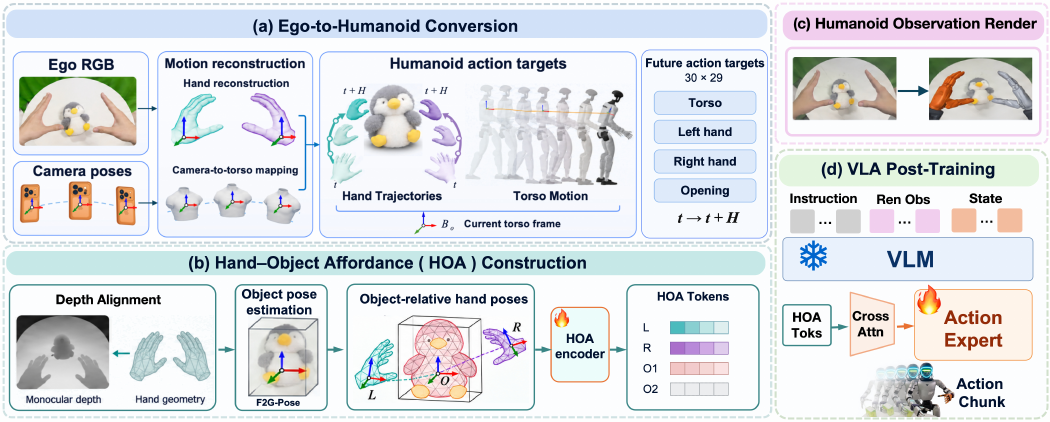}
\caption{\textbf{Pipeline of \egoloaf{}.} \textbf{(a)} Hand reconstruction and camera-to-torso mapping produce future torso and hand action targets in the current torso frame. \textbf{(b)} Monocular depth is scale-aligned with reconstructed hand geometry for object-pose estimation. Current entity poses and object-relative hand poses are encoded as hand--object affordance (HOA) tokens. \textbf{(c)} Human hands and forearms are replaced with rendered robot counterparts. \textbf{(d)} With the VLM frozen, the trainable action expert predicts action chunks from rendered observations, instructions, and current state, conditioned on HOA tokens through cross-attention. The targets in (a) provide action supervision.}
\label{fig:overview}
\end{figure}

\egoloaf{} converts egocentric demonstrations into humanoid action targets, rendered training observations, and current hand--object affordances (Fig.~\ref{fig:overview}). We use the converted data to post-train a VLA policy, with affordances conditioning its action expert and the vision-language backbone frozen. At deployment, robot measurements provide the same affordance representation, and a whole-body controller executes the predicted torso and hand motion.

\subsection{Egocentric-to-humanoid data construction}
\label{sec:actions}
Each recording contains RGB images $I_t$, camera intrinsics $K_t$, timestamps, and camera poses in a recording-local world frame $W$, without wearable motion trackers. The capture setup is described in Sec.~\ref{sec:dataset}. We write $\trans{A}{B}$ for the pose of frame $B$ expressed in frame $A$.

\textbf{Motion reconstruction.} We calibrate the camera-to-torso transform $M=\trans{B}{C}$ on the target humanoid. The recorded camera trajectory defines a virtual torso trajectory as $\trans{W}{B_t}=\trans{W}{C_t}M^{-1}$. HaWoR \citep{hawor} reconstructs MANO hand meshes and keypoints \citep{mano} in the camera frame $C_t$. We define a hand tool center point (TCP) at the midpoint of the thumb and index fingertips, with orientation determined by wrist and proximal palm landmarks. The corresponding left and right TCP frames are $H_t^L,H_t^R$. Composing each camera-frame TCP pose with the recorded camera pose gives its world-frame trajectory. This representation links human and robot hand motion without matching their joint configurations. We also derive binary opening states $g_t^h$ from palm-normalized thumb--index separation, using temporal filtering and fixed per-hand calibration bounds; $h\in\{L,R\}$ and 1 denotes open.

\textbf{Action targets.} At observation time $t_o$, we use the current torso position and orientation as a fixed reference $B_o=B_{t_o}$ for subsequent torso and hand motion. Each world-frame pose is expressed in this reference as
\begin{equation}
 \trans{B_o}{X_t}=(\trans{W}{B_o})^{-1}\trans{W}{X_t},
 \quad X\in\{B,H^L,H^R\}.
 \label{eq:fixedframe}
\end{equation}
The reference stays fixed throughout each chunk, so torso motion does not change the spatial meaning of the hand targets. Each pose is encoded by three normalized translation coordinates and the first two rotation-matrix columns \citep{rotation6d}. At future time $t_j$, let $\mathbf b_{o,j}$ denote the torso's nine-dimensional pose vector and $\mathbf h^L_{o,j},\mathbf h^R_{o,j}$ the corresponding hand vectors. The action target is
\begin{equation}
 \mathbf a_{o,j}=\big[\mathbf b_{o,j};\,\mathbf h^L_{o,j};\,
 \mathbf h^R_{o,j};\,g^L_{t_j};\,g^R_{t_j}\big]\in\mathbb R^{29}.
 \label{eq:action}
\end{equation}
Each chunk contains 30 targets, indexed by $j=1,\ldots,30$. The current torso and TCP poses provide the pose state in the same reference; current opening states are supplied separately.

\textbf{Rendered observations.} We fit robot-hand meshes to the reconstructed hands, erase visible human hands and forearms, and fill the removed regions with LaMa \citep{lama}. Visibility labels from the original hand and object masks are transferred to the robot meshes before compositing, preserving hand--object occlusions. Each rendered frame $\widetilde I_{t_o}$ is paired with the current state and future action targets from the same recording. The supplementary material gives the TCP construction, opening calibration, mesh fitting, and visibility-transfer details.

\subsection{Hand--object affordance construction}
\label{sec:ict}
The action targets specify the motion to execute; hand--object affordances describe the configuration from which that motion starts. We encode these affordances with current entity poses and the position and orientation of each hand relative to each entity \citep{humanego}. For current entity $E_k$ and hand $h$, the relative pose is
\begin{equation}
 \trans{E_k}{H^h}
 =(\trans{B_o}{E_k})^{-1}\trans{B_o}{H^h}.
 \label{eq:handobject}
\end{equation}
These relative poses are invariant to a common rigid change of reference. We use four entity slots: the two hand TCPs, the manipulated object, and an optional second object. Each entity's 29-dimensional vector contains its type, its pose in $B_o$, both hand poses in the entity frame, and an opening indicator. The four vectors form a $4\times29$ affordance input for the current observation. The opening indicator stores the open/closed state for hand rows and $-1$ for object rows. Entity and feature masks mark missing or invalid measurements. The supplementary material gives the complete feature definition.

\textbf{Estimation from human demonstrations.} We reuse the reconstructed hand poses and opening states. The metric variant of Depth Anything V2 \citep{depthanythingv2} provides monocular depth, but its scale prior can differ from that of the hand estimator. We align the depth map using the median ratio of reconstructed hand-surface depth to predicted depth at corresponding visible pixels, with temporal smoothing. F2G-Pose \citep{f2g} then estimates object poses from RGB, object masks, and the aligned depth. Camera poses map the objects into $B_o$ alongside the hands, allowing us to compute Eq.~\ref{eq:handobject}. All geometry is estimated from the original recordings; inpainted pixels are used only in rendered policy observations.

\subsection{Affordance-conditioned post-training and deployment}
\label{sec:execution}
\textbf{Affordance conditioning.} We post-train \pifive{} \citep{pi05} with rendered images, task instructions, current pose and opening states, and current affordances. An encoder projects each entity row and its feature-validity mask into an affordance token; we collect these tokens in $\mathbf Z$. Before selected action-expert layers, the incoming hidden states $\mathbf H_\ell$ receive a residual cross-attention update:
\begin{equation}
 \widetilde{\mathbf H}_\ell=\mathbf H_\ell+
 \operatorname{CA}_\ell(\mathbf H_\ell,\mathbf Z;\mathbf m).
 \label{eq:affordancefusion}
\end{equation}
Here $\mathbf m$ masks invalid entities, and $\operatorname{CA}_\ell$ includes query normalization and learned projections. Expert hidden states provide the queries; affordance tokens provide the keys and values. Its output is zero when no entities are valid. The updated states enter the action-expert layer, giving action generation direct access to current hand--object relations while the vision-language backbone remains frozen.

\textbf{Post-training.} We update the action expert, action input/output projections, and affordance-conditioning modules. The action flow-matching loss supervises the 30-target chunks in Eq.~\ref{eq:action}, excluding invalid target components. Affordances enter as current conditions; only future actions are prediction targets.

\textbf{Online affordances and execution.} At deployment, robot forward kinematics and hand feedback provide TCP poses and opening states. F2G-Pose uses head-camera RGB-D to estimate object poses. We encode these measurements using the same affordance representation as during post-training. The policy uses robot RGB, task instructions, current state, and these affordances to predict the next action chunk. Fixed tool transforms convert predicted TCP poses to wrist targets. A frozen ScaleBFM controller \citep{scalebfm}, fine-tuned by KODA \citep{koda}, tracks these targets together with the torso motion. Hand opening is commanded separately. Real-time chunking \citep{rtc} and segment blending connect asynchronous policy predictions to continuous control. We apply spatial feedback to compensate for torso-tracking errors during execution. The supplementary material details this correction, tool transforms, execution timing, and the remaining implementation settings.

\section{The \egoloaf{} dataset}
\label{sec:dataset}

We release the \egoloaf{} dataset for humanoid loco-manipulation, comprising \textbf{128,400 complete, independently recorded demonstrations across 1,284 tasks}. Each task has 100 demonstrations. After quality filtering, the recordings total \textbf{1,000 hours}.

\begin{figure}[t]
\centering
\includegraphics[trim={35bp 0bp 19.5bp 0bp},clip,width=\textwidth]{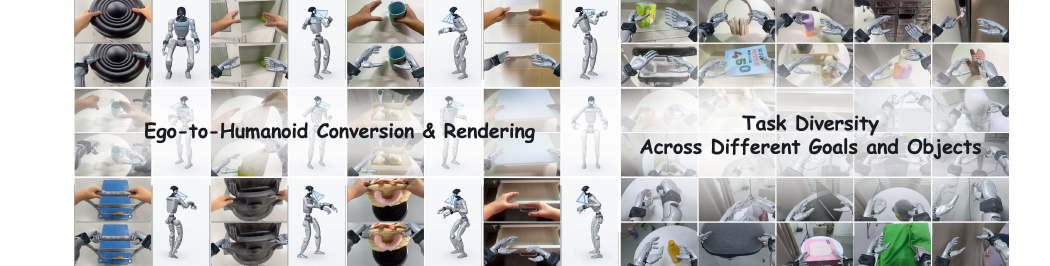}
\caption{\textbf{The \egoloaf{} dataset.} Left: paired original egocentric observations and observations with rendered G1 arms and O6 hands, alongside G1 poses with camera frusta. Right: rendered observations illustrate task diversity across manipulation goals and objects.}
\label{fig:dataset}
\end{figure}

\subsection{Collection and task coverage}
We collected demonstrations with a chest-mounted iPhone, without wearable motion trackers. Camera height and pitch approximately match those of the G1 head camera. The Stera application from MobileEgo Anywhere \citep{mobileego} records RGB images, timestamps, camera intrinsics, and camera trajectories. This setup leaves both hands free and requires no robot during collection.

Tasks span different manipulation goals and objects. Examples include object placement, pouring, and interactions with containers and cabinets. The collection includes coordinated two-hand manipulation and body movements such as walking, turning, and bending. Recordings preserve any transitions between body repositioning and object manipulation. Figure~\ref{fig:dataset} shows representative tasks and paired original and rendered observations.

\subsection{Annotations and data organization}
Each demonstration includes the original RGB observations, a task instruction, estimated torso and bilateral hand trajectories, and binary hand-opening states. The torso and hand trajectories include both position and orientation over time. Object poses and precomputed hand--object affordances are provided for all demonstrations. The affordances encode the current hand poses relative to the objects, following Sec.~\ref{sec:ict}.

We also provide rendered observations that replace human arms and hands with G1 arms and O6 dexterous hands, matching the configuration used in our real-world experiments. Original and rendered frames share timestamps, motion targets, and affordance annotations. All motion and affordance annotations are computed from the original recordings.

The task instruction applies to the complete demonstration, while poses, opening states, and affordances are associated with individual observations. Training windows pair the current observation and hand--object affordances with subsequent torso and hand motion as action supervision. Continuous trajectories support sampling action targets at different prediction horizons without repeating motion reconstruction.

We checked each demonstration for recording completeness, camera-trajectory quality, arm and hand rendering quality, and task completion. Entire demonstrations that failed these checks were discarded.

For each of the four experimental tasks, post-training uses 100 demonstrations from the released dataset and 50 collected separately (Sec.~\ref{sec:experiments}). Checkpoint selection uses separate held-out recordings. All windows from these recordings are excluded from post-training.

\section{Experiments}
\label{sec:experiments}
We evaluate whether \egoloaf{} learns effective humanoid loco-manipulation from egocentric human task demonstrations. We compare it with a teleoperation-based baseline under matched training-frame budgets (Sec.~\ref{sec:overall_performance}), then examine how hand--object affordance conditioning benefits policies trained from human and robot demonstrations (Sec.~\ref{sec:component_ablations}).

\begin{figure}[t]
\centering
\includegraphics[width=\textwidth]{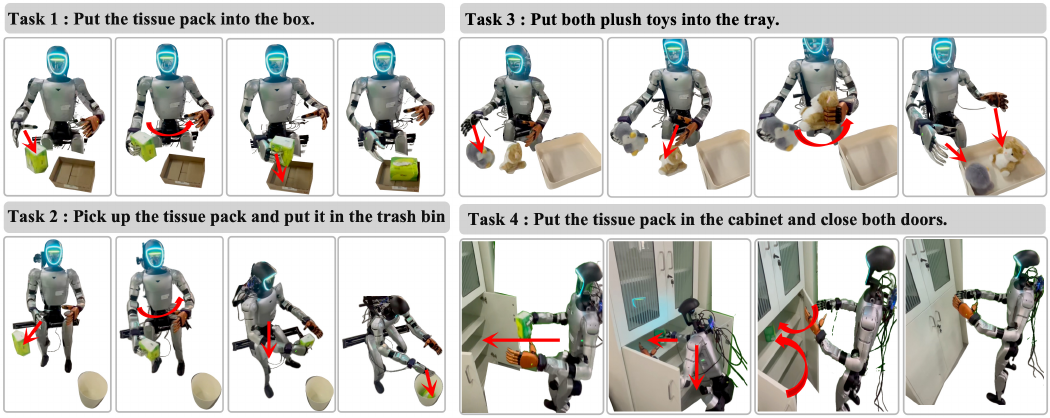}
\caption{\textbf{Real-world humanoid loco-manipulation.} \textbf{Task~1} tests grasping, torso turning, and precise placement of a tissue pack into a box. \textbf{Task~2} combines turning, squatting, and bending to place a tissue pack into a low bin. \textbf{Task~3} tests bimanual coordination and turning when placing both plush toys into a tray. \textbf{Task~4} combines walking, half-squatting, placement inside a cabinet, and closing both doors. Each evaluation batch contains 20 trials per task for each policy.}
\label{fig:experiment_tasks}
\end{figure}

\subsection{Experimental setup}
\label{sec:experimental_setup}
\textbf{Robot platform.} We use a Unitree G1 humanoid with LinkerHand O6 dexterous hands and a head-mounted Intel RealSense D435 RGB-D camera. We calibrate the camera-to-torso transform and construct online affordances following Secs.~\ref{sec:ict} and~\ref{sec:execution}. Grounding DINO \citep{groundingdino} initializes or recovers object regions, while mask propagation and SAM refinement \citep{sam} support subsequent pose estimates.

\textbf{Execution settings.} All policy comparisons share the same frozen whole-body controller and execution settings, including spatial feedback. Only the explicitly labeled feedback ablations in the supplementary material disable this additional correction.

\begin{figure}[t]
\centering
{\footnotesize
\textcolor[HTML]{2878A6}{\rule{1em}{0.65em}}\enspace EgoLoAF (ego data)
\qquad
\textcolor[HTML]{D9824C}{\rule{1em}{0.65em}}\enspace Teleoperation baseline (robot data)\par}
\smallskip
\begin{subfigure}[t]{0.485\textwidth}
  \vspace{0pt}
  \centering
  \includegraphics[width=\linewidth]{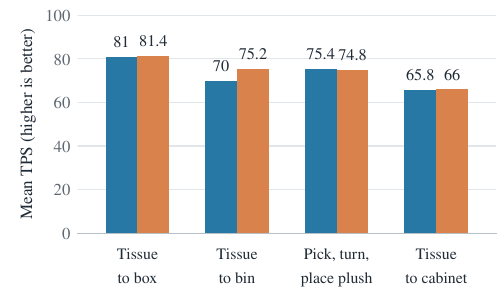}
  \caption{Task performance.}
  \label{fig:overall_tps}
\end{subfigure}\hfill
\begin{subfigure}[t]{0.485\textwidth}
  \vspace{0pt}
  \centering
  \includegraphics[width=\linewidth]{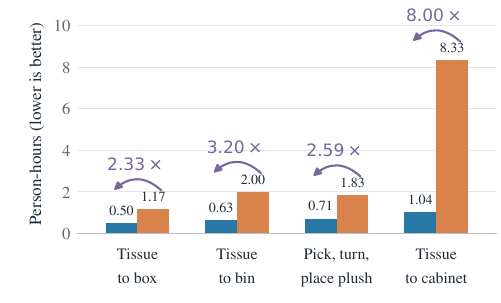}
  \caption{Data collection effort.}
  \label{fig:overall_effort}
\end{subfigure}
\caption{\textbf{Comparison with teleoperation.}
\subref{fig:overall_tps} Mean TPS (0--100, $\uparrow$), aggregated across evaluation batches of 20 trials per task. EgoLoAF uses 150 egocentric human demonstrations per task; robot training data match their frame counts. Both methods use spatial feedback; the teleoperation baseline has no affordance input.
\subref{fig:overall_effort} Acquisition person-hours at matched training-frame budgets ($\downarrow$): one collector for ego data and two for teleoperation. Arrow labels give the teleoperation/ego person-hour ratio for each task.}
\label{fig:overall_comparison}
\end{figure}

\begin{figure}[t]
\centering
{\footnotesize
\begin{tabular}{@{}ll@{}}
\textcolor[HTML]{2878A6}{\rule{1em}{0.65em}}\enspace With affordances &
\textcolor[HTML]{D9824C}{\rule{1em}{0.65em}}\enspace Without affordances
\end{tabular}\par}
\smallskip
\includegraphics[width=0.485\linewidth]{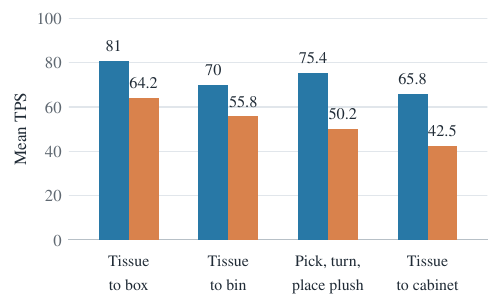}
\caption{\textbf{Effect of affordances on human-data policies.}
Mean TPS (0--100, $\uparrow$), aggregated across evaluation batches of 20 trials per task/configuration, using the same 150 egocentric human demonstrations per task. Policies are trained separately with and without affordances; both use spatial feedback. All four tasks use affordances in the conditioned policy.}
\label{fig:human_ablation}
\end{figure}

\textbf{Real-world tasks.} Following the evaluation practice of prior studies \citep{egohumanoid,egohumanoidv2}, we select four real-world tasks to evaluate increasingly complex combinations of hand manipulation and body motion. These tasks range from precise placement to bimanual manipulation and long-horizon sequences involving turning, walking, bending, and squatting (Fig.~\ref{fig:experiment_tasks}):
\begin{enumerate}
    \item \textbf{Tissue-to-Box.} This precision task requires the robot to grasp a tissue pack on the table, turn, and place it precisely into a box.
    \item \textbf{Tissue-to-Bin.} The robot grasps a tissue pack, then turns, squats, and bends to place it into a low bin, combining fine hand manipulation with whole-body motion.
    \item \textbf{Pick-Turn-Place.} The robot sequentially grasps two plush toys on the table, one in each hand, then turns and places both into a tray, combining coordinated two-hand manipulation with torso motion.
    \item \textbf{Tissue-to-Cabinet.} The robot grasps a tissue pack from the table, turns and walks toward a cabinet, then half-squats to place it inside and closes both cabinet doors. This long-horizon task combines sequential hand manipulation with turning, walking, and squatting.
\end{enumerate}

\textbf{Evaluation protocol.} For a fair comparison, each evaluation batch contains 20 trials per task for each policy checkpoint under the same protocol with randomized initial scene configurations. Following prior progress-based evaluations \citep{hilumi,egohumanoidv2,pi05}, we report the mean Task Progress Score (TPS), aggregated across multiple evaluation batches. Each trial's TPS is the sum of the points awarded for completed subtasks, with a maximum of 100. Table~\ref{tab:tps_criteria} lists the subtasks and point allocations, fixed before testing.
\begin{table}[t]
\centering
\caption{\textbf{Task Progress Score criteria.} Each task totals 100 points. For repeated subtasks, scores are written as points per object $\times$ object count.}
\label{tab:tps_criteria}
\small
\setlength{\tabcolsep}{3pt}
\renewcommand{\arraystretch}{1.08}
\begin{tabular*}{\linewidth}{@{\extracolsep{\fill}}llr@{}}
\toprule
Task & Subtask & Score\\
\midrule
\multirow{3}{*}{Tissue-to-Box} & Grasp the tissue pack & $+30$\\
 & Move tissue above the box & $+30$\\
 & Release tissue into the box & $+40$\\
\midrule
\multirow{4}{*}{Tissue-to-Bin} & Grasp the tissue pack & $+25$\\
 & Turn toward the bin & $+25$\\
 & Squat down & $+25$\\
 & Place tissue in the bin & $+25$\\
\midrule
\multirow{3}{*}{Pick-Turn-Place} & Grasp a plush toy & $+20\times 2$\\
 & Turn and move toys over the tray & $+20$\\
 & Release a toy into the tray & $+20\times 2$\\
\midrule
\multirow{6}{*}{Tissue-to-Cabinet} & Grasp the tissue pack & $+15$\\
 & Turn toward the cabinet & $+15$\\
 & Walk to the cabinet & $+15$\\
 & Half-squat & $+15$\\
 & Place tissue in the cabinet & $+20$\\
 & Close a cabinet door & $+10\times 2$\\
\bottomrule
\end{tabular*}
\end{table}

\textbf{Training data.} For each task, we post-train a separate \pifive{} policy \citep{pi05} on 150 egocentric human demonstrations. Validation recordings are held out separately for checkpoint selection.

\subsection{Comparison with teleoperation}
\label{sec:overall_performance}
We first assess whether a policy post-trained solely on egocentric human task demonstrations can learn effective robot actions. The teleoperation baseline is trained on demonstrations collected with the same G1, recording head-camera images and body and hand states, with the same number of training frames as the ego data for each task. Both methods use the action representation and policy/controller architecture in Secs.~\ref{sec:actions} and~\ref{sec:execution}, with matched optimization budgets. \egoloaf{} additionally uses affordances on all four tasks, while the teleoperation baseline uses no affordance input.

\egoloaf{} achieves a four-task mean TPS of 73.1, compared with 74.4 for teleoperation (Fig.~\ref{fig:overall_comparison}\subref{fig:overall_tps}). It trails the baseline by 0.4 points on box and 5.2 points on bin, and exceeds it by 0.6 points on plush. On the long-horizon Tissue-to-Cabinet task, it achieves 65.8 TPS compared with 66.0 for teleoperation. These results show that egocentric human task demonstrations provide effective supervision for humanoid loco-manipulation, including a sequence that combines grasping, turning, walking, squatting, placement, and door closing. The comparison evaluates the complete pipelines; Sec.~\ref{sec:component_ablations} isolates the effect of affordance conditioning within each data source.

\label{sec:efficiency}
We also evaluate data-collection efficiency using person-hours \citep{umi}, defined as the total working time, in hours, summed across all participants. For matched training-frame budgets, we count the time spent on setup, demonstration collection, failed attempts, and resets. Automated processing time that requires no human involvement is excluded from person-hours. Egocentric collection uses one person, while teleoperation uses two.

Figure~\ref{fig:overall_comparison}\subref{fig:overall_effort} shows a mean \textbf{$4.03\times$ improvement in acquisition efficiency}. Among the four tasks, Tissue-to-Cabinet shows the largest gain ($8.00\times$). Resetting the humanoid and task scene is time-consuming during teleoperation, especially for long-horizon tasks, which contributes to the larger efficiency gap on this task. Collection details are provided in the supplementary material.

\subsection{Ablation: effect of hand--object affordances}
\label{sec:component_ablations}
We next isolate the effect of hand--object affordances within each data source, using policies trained from human and robot demonstrations.

\subsubsection{Human-data policies}
\label{sec:ablations}
Policies with and without affordance inputs use identical training frames, targets, masks, optimization budgets, and execution settings. Figure~\ref{fig:human_ablation} compares these policies on all four tasks, in the same order as the main comparison.

Adding affordances raises the four-task mean TPS from 53.2 to 73.1, a gain of 19.9 points.
TPS increases from 64.2 to 81.0 on Tissue-to-Box and from 55.8 to 70.0 on Tissue-to-Bin.
Larger gains occur on Pick-Turn-Place, from 50.2 to 75.4 (+25.2), and Tissue-to-Cabinet, from 42.5 to 65.8 (+23.3).
The improvement narrows the gap to the affordance-free teleoperation baseline (74.4 TPS) from 21.2 to 1.3 points.

Pick-Turn-Place and Tissue-to-Cabinet require multiple object interactions combined with body repositioning.
In observed executions, affordance conditioning helps the robot adjust reaching and placement to the current hand--object relationship.
This improves contact and coordination between hand actions and body motion across successive task stages.
These observations explain why pairing converted motion targets with current affordance conditions makes egocentric supervision more effective for robot manipulation.

\subsubsection{Robot-data policies}
\label{sec:robot_affordances}
Affordances also improve policies trained from robot demonstrations (Table~\ref{tab:ict_results}).
With robot training data, optimization budgets, and execution settings held fixed, the four-task mean TPS increases from 74.4 to 86.3, a gain of 11.9 points.
TPS rises from 81.4 to 90.0 on Tissue-to-Box, from 75.2 to 80.0 on Tissue-to-Bin, and from 66.0 to 75.0 on Tissue-to-Cabinet.
The largest gain again occurs on Pick-Turn-Place, from 74.8 to 100.0 (+25.2).

The improvement on robot data shows that HOA remains useful when training and deployment use the same embodiment.
Even with robot motion targets, the policy must determine how its hands should approach and contact objects from the current observation.
Providing the hand--object relationship explicitly helps guide these actions, complementing trajectory supervision for both human-data and robot-data policies.
\begin{table}[!h]
\centering
\caption{\textbf{Affordances for robot-data policies.} Mean TPS (0--100, $\uparrow$), with 20 trials per task per evaluation batch. Both policies use identical robot training data and spatial feedback. Mean weights all four tasks equally; Cabinet denotes Tissue-to-Cabinet.}
\label{tab:ict_results}
\small
\setlength{\tabcolsep}{3pt}
\begin{tabular}{@{}lccccc@{}}
\toprule
Robot policy & Box & Bin & Plush & Cabinet & Mean\\
\midrule
w/o affordances & 81.4 & 75.2 & 74.8 & 66.0 & 74.4\\
with affordances & \textbf{90.0} & \textbf{80.0} & \textbf{100.0} & \textbf{75.0} & \textbf{86.3}\\
\bottomrule
\end{tabular}
\end{table}

\section{Conclusion}
\label{sec:conclusion}
We presented \egoloaf{}, a paradigm that converts egocentric human demonstrations into humanoid loco-manipulation training data without wearable motion trackers.
Each recording provides coordinated torso and hand motion as action supervision, rendered robot-view observations, and current hand--object affordances.
We post-train the VLA entirely on the converted data, without real-robot teleoperation demonstrations.
We inject the affordances into the action expert through cross-attention while keeping the vision-language backbone frozen.
We also release 128,400 egocentric demonstration episodes spanning 1,000 hours and 1,284 tasks, together with their converted humanoid training data.
Across four real-world tasks, \egoloaf{} nearly matches the teleoperation-trained baseline (73.1 vs.\ 74.4 mean TPS), while more than quadrupling data-collection efficiency on average ($4.03\times$).

\section*{LLM Usage Statement}
We used OpenAI Codex to assist with literature research,
language polishing and translation of the manuscript,
and the organization of supplementary materials.
The authors reviewed and verified all AI-assisted content,
including the cited references, and take full responsibility
for the final manuscript and supplementary materials.

\FloatBarrier
\bibliographystyle{plainnat}
\bibliography{main}

\clearpage
\pdfbookmark[0]{Appendix}{appendix}
\beginappendix
\setcounter{figure}{0}
\setcounter{table}{0}
\setcounter{equation}{0}
\renewcommand{\thefigure}{S\arabic{figure}}
\renewcommand{\thetable}{S\arabic{table}}
\renewcommand{\theequation}{S\arabic{equation}}
\renewcommand{\theHfigure}{appendix.\arabic{figure}}
\renewcommand{\theHtable}{appendix.\arabic{table}}
\renewcommand{\theHequation}{appendix.\arabic{equation}}
This appendix details data acquisition and conversion (Sec.~\ref{sec:sacquisition}), affordance-conditioned training (Sec.~\ref{sec:straining}), robot execution (Sec.~\ref{sec:sexecution}), evaluation protocols and additional results (Sec.~\ref{sec:sprotocol}), and dataset coverage and organization (Sec.~\ref{sec:srelease}).
\section{Data acquisition and action construction}
\label{sec:sacquisition}
We collect RGB images, camera intrinsics, timestamps, and camera poses using a chest-mounted iPhone running Stera from MobileEgo Anywhere~\citep{mobileego}. Camera height and pitch approximately match the robot head camera. HaWoR~\citep{hawor} reconstructs camera-frame MANO~\citep{mano} hand geometry. Observations and motion trajectories are aligned using recorded timestamps, with training samples organized under a 30\,Hz convention.

\subsection{Camera, torso, and hand trajectories}
Let $M=\trans{B}{C}$ be the calibrated camera-to-torso transform on the target humanoid. The recorded camera trajectory defines the virtual torso trajectory as
\begin{equation}
 \trans{W}{B_t}=\trans{W}{C_t}M^{-1}.
 \label{eq:virtualtorso}
\end{equation}
At observation time $t_o$, we fix $B_o=B_{t_o}$ and express subsequent poses in that frame:
\begin{equation}
 \trans{B_o}{X_t}=(\trans{W}{B_o})^{-1}\trans{W}{X_t},
 \quad X\in\{B,H^L,H^R\}.
 \label{eq:sfixedframe}
\end{equation}
For a camera-frame hand TCP pose, the complete mapping is
\begin{equation}
 \trans{B_o}{H_t^h}
 =(\trans{W}{B_o})^{-1}\trans{W}{C_t}\trans{C_t}{H_t^h}.
 \label{eq:handtcp}
\end{equation}
The reference $B_o$ stays fixed throughout each action chunk. Each pose uses three translation coordinates and the first two rotation-matrix columns; we denote its normalized encoding by $\bar\pose$. The torso and two hand poses, together with two opening states, form the 29-dimensional action defined in the main paper.

\subsection{Hand tool frame}
\label{sec:stcp}
\begin{figure}[t]
\centering
\includegraphics[width=0.65\linewidth]{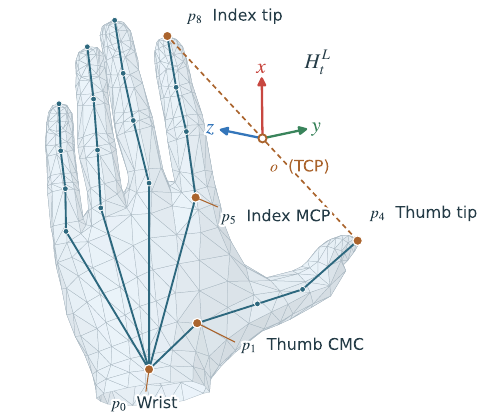}

\caption{\textbf{Hand keypoints and TCP frame.} Dorsal view of a HaWoR-reconstructed MANO mesh with all 21 keypoints and five finger chains. The five labeled landmarks define the TCP frame $H_t^L$. Its origin $o$ is the fingertip midpoint; red, green, and blue arrows show the projected $x,y,z$ axes.}
\label{fig:shandkeypoints}
\end{figure}

Let $p_i$ be the 21 predicted hand keypoints (Fig.~\ref{fig:shandkeypoints}). The TCP uses the wrist $p_0$, thumb carpometacarpal (CMC) joint $p_1$, index metacarpophalangeal (MCP) joint $p_5$, and thumb and index fingertips $p_4,p_8$:
\begin{equation}
\begin{aligned}
 o&=(p_4+p_8)/2,\qquad x=\frac{p_5-p_1}{\|p_5-p_1\|},\\
 u&=(p_1+p_5)/2-p_0,\\
 \tilde y&=\frac{u-x(x^\top u)}{\|u-x(x^\top u)\|},\\
 z&=\frac{x\times\tilde y}{\|x\times\tilde y\|},\qquad y=z\times x,\\
 T_H&=\begin{bmatrix}x&y&z&o\\0&0&0&1\end{bmatrix}.
\end{aligned}
\label{eq:stcpframe}
\end{equation}
This follows the virtual-gripper construction in HumanEgo~\citep{humanego}, with the selected proximal thumb reference. We temporally smooth valid hand trajectories in world coordinates and construct orthonormal TCP frames. Missing or unreliable observations are masked out of action supervision. The robot-side tool mapping is given in Sec.~\ref{sec:sexecution}.

\subsection{Automatic hand-opening labels}
\label{sec:sopening}
We derive binary hand-opening labels from the thumb--index fingertip distance normalized by palm length. For hand $h$, let $p_9$ denote the middle-finger MCP joint. We temporally smooth the aperture ratio $r_t^h$ to obtain $\widetilde r_t^h$, then normalize it using fixed per-hand calibration bounds $r_{\min}^h$ and $r_{\max}^h$, shared across episodes:
\begin{equation}
\begin{aligned}
 r_t^h&=\frac{\|p_{4,t}^h-p_{8,t}^h\|_2}{\|p_{9,t}^h-p_{0,t}^h\|_2},\\
 u_t^h&=\operatorname{clip}\!\left(
 \frac{\widetilde r_t^h-r_{\min}^h}{r_{\max}^h-r_{\min}^h},0,1\right),\\
 g_t^h&=\mathbf{1}[u_t^h\geq0.5].
\end{aligned}
\label{eq:aperture}
\end{equation}
Here, $g_t^h=1$ denotes an open hand and $g_t^h=0$ a closed hand.

\subsection{Rendered robot observations}
\label{sec:srendering}
We generate robot-view observations by replacing human hands and forearms with O6 hands and G1 arms. We fit the robot-hand meshes to the reconstructed MANO geometry through landmark alignment and pose optimization under the robot's joint constraints. The fitted configurations determine the robot appearance in each frame.

To preserve hand--object occlusions, we transfer surface visibility from the reconstructed human hands to the robot meshes. MobileSAM~\citep{mobilesam} provides visible hand and object masks from the original RGB frame. We project the MANO meshes into the image and render their depth to identify the frontmost hand surfaces. Frontmost surfaces supported by the hand masks receive visible labels, while projected hand surfaces covered by the visible object mask receive object-occlusion labels. We establish correspondence between MANO and O6 surfaces using the local structure of the palm and individual fingers. Each rendered robot-hand fragment then retrieves its visibility label from the corresponding location on the posed MANO surface. A fragment is hidden when it carries an object-occlusion label and its image projection overlaps the visible object mask. A Z-buffer resolves occlusions within and between the rendered robot hands.

For image compositing, we remove the visible human hands and forearms and use LaMa~\citep{lama} to complete the exposed regions. We then composite the robot rendering onto the completed image according to the transferred visibility labels, retaining the object in front of the occluded robot-hand regions. Action targets and affordances are computed from the original recording and paired with the rendered observations through their timestamps.

\section{Affordance construction and policy training}
\label{sec:straining}
\subsection{Depth calibration}
\label{sec:sdepth}
We estimate a depth scale factor to align the monocular depth map $D_t$ from Depth Anything V2~\citep{depthanythingv2} with reconstructed hand geometry. Let $\mathcal V_t^h$ contain reliable visible hand-surface vertices, excluding object-occluded regions. For a hand vertex $v$ with camera-frame depth $z_v$ and image projection $\pi_{K_t}(v)$, the per-hand scale factor is
\begin{equation}
 s_t^h=\med_{v\in\mathcal V_t^h}
 \frac{z_v}{D_t(\pi_{K_t}(v))}.
 \label{eq:handscale}
\end{equation}
Valid scale factors from both hands are combined by their geometric mean and temporally smoothed in log space. We multiply the predicted depth map by the resulting scale factor before estimating object poses. F2G-Pose~\citep{f2g} estimates object poses from the aligned depth, original RGB, and object masks. Camera poses place these estimates in the same reference as the hands. Offline annotation may use neighboring frames for smoothing; at deployment, object poses are estimated from the robot's current RGB-D observations.

\subsection{Affordance representation}
\label{sec:s_affordance_representation}
For a current entity $E_k$, the affordance vector is
\begin{equation}
\begin{split}
 z_k=\big[&\tau_k;\,\pose(\trans{B_o}{E_k});\,
 \pose(\trans{E_k}{H^L});\,\\
 &\pose(\trans{E_k}{H^R});\,g_k\big]\in\mathbb R^{29}.
\end{split}
 \label{eq:ict}
\end{equation}
Table~\ref{tab:ictfields} gives its components. Four slots represent the two hand TCPs and up to two objects, yielding a $4\times29$ input at the current observation. Entity-local hand relations are invariant to a common rigid change of reference. Feature masks mark unavailable components, and entity masks exclude invalid entities from attention.

\begin{table}[htbp]
\centering
\caption{\textbf{Affordance features per entity.} The four entity slots share this representation.}
\label{tab:ictfields}
\small
\setlength{\tabcolsep}{3pt}
\begin{tabular}{@{}l>{\raggedright\arraybackslash}p{4.2cm}r@{}}
\toprule
Feature & Meaning & Dim.\\
\midrule
Entity type & Left/right hand, object 1 or 2 & 1\\
Entity pose & Pose in the current torso frame & 9\\
Left-hand relation & Left TCP pose in entity frame & 9\\
Right-hand relation & Right TCP pose in entity frame & 9\\
Opening indicator & Hand opening; $-1$ for an object & 1\\
\bottomrule
\end{tabular}
\end{table}
\FloatBarrier

\subsection{Affordance fusion and action objective}
The encoder separately projects each normalized entity vector and its feature-validity mask, sharing projection weights across slots. We sum these projections, add a learned slot embedding, and apply layer normalization to form the affordance token. Residual cross-attention follows Sec.~\ref{sec:execution} of the main paper and precedes every third action-expert block. Table~\ref{tab:smodelconfig} summarizes the configuration.

\begin{table}[t]
\centering
\caption{\textbf{Affordance encoder and fusion configuration.}}
\label{tab:smodelconfig}
\small
\begin{tabular*}{\linewidth}{@{\extracolsep{\fill}}ll@{}}
\toprule
Component & Configuration\\
\midrule
Entity-feature projection & Linear, $29\rightarrow256$\\
Feature-mask projection & Linear, $29\rightarrow256$\\
Slot embeddings & Learned, $4\times256$\\
Token normalization & LayerNorm\\
Cross-attention heads & 4\\
Fusion & Residual cross-attention\\
Residual output projection & Zero initialization\\
\bottomrule
\end{tabular*}
\end{table}

Let $A\in\mathbb R^{30\times29}$ be an action chunk, $M_A$ its component-validity mask, $\epsilon$ Gaussian noise, and $u\in[0,1]$ the sampled noise time. We use the action flow-matching objective
\begin{equation}
\begin{aligned}
 X_u&=(1-u)A+u\epsilon,\qquad v^\star=\epsilon-A,\\
 \mathcal L_{\mathrm{action}}&=
 \frac{\sum M_A\odot\bigl(v_\theta(X_u,u\mid c)-v^\star\bigr)^2}
 {\max(\sum M_A,1)}.
\end{aligned}
 \label{eq:sflowloss}
\end{equation}
The conditioning $c$ contains the current image, instruction, pose and opening states, and affordances when enabled. Only future actions are supervised, with the loss averaged over valid components of the normalized action representation.

\subsection{Post-training}
Each task policy initializes from \pifive{}~\citep{pi05}. We freeze the vision-language backbone and update the action expert, action input/output projections, and affordance-conditioning modules. Human-data policies use rendered observations, while robot-data policies use recorded head-camera observations. The affordance-free policy omits the adapter and is trained separately. Training-data controls are specified in Sec.~\ref{sec:sprotocol}.

\section{Real-robot deployment}
\label{sec:sexecution}
We deploy on a Unitree G1 with LinkerHand O6 hands and a head-mounted Intel RealSense D435 RGB-D camera. We use an NVIDIA GeForce RTX 5090 GPU for inference.

\subsection{Online perception and affordance construction}
Grounding DINO~\citep{groundingdino} initializes task-object regions and reacquires lost targets. Optical-flow mask propagation and periodic SAM refinement~\citep{sam} update these regions between detections. F2G-Pose~\citep{f2g} takes RGB, measured depth, object masks, and camera intrinsics to estimate camera-frame object poses. The calibrated camera-to-torso transform maps them into the observation-time torso frame $B_o$.

Robot joint feedback and forward kinematics provide wrist poses, which the fixed tool transforms map to TCP poses; hand feedback provides opening states. We associate these measurements with RGB-D frames using nearby host receive timestamps. The resulting hand and object measurements populate the four-slot affordance representation in Sec.~\ref{sec:s_affordance_representation}, with masks for unavailable features. Policy inference consumes the RGB image, instruction, current state, and these affordances.

\subsection{Asynchronous policy execution}
Each prediction uses 10 flow-matching integration steps to generate 30 future 29-dimensional actions. The executor interpolates the timestamped targets while the policy computes the next chunk asynchronously. New plans are aligned in time with the ongoing trajectory and joined by segment blending. We use real-time chunking (RTC)~\citep{rtc} to condition generation of the next chunk on the already committed action prefix.

Let $G_h$ be robot wrist frame $h$ and $F_h=\trans{G_h}{H_h^{\mathrm{open}}}$ the fixed transform to its nominal open-hand TCP. A predicted TCP target becomes a wrist target through
\begin{equation}
 \trans{B_o}{G_h^\star}=\trans{B_o}{H_h^\star}F_h^{-1}.
 \label{eq:toolmap}
\end{equation}
Predictions, feedback, and affordances share this tool convention. The frozen Torso-UMI-3 checkpoint of ScaleBFM~\citep{scalebfm}, fine-tuned by KODA~\citep{koda}, tracks the torso and wrist targets at 50\,Hz. Hand opening is commanded separately, with hysteresis and temporal filtering. During policy inference, the controller continues tracking the current plan; if the plan is exhausted, the reference decelerates to a hold.

\subsection{Spatial feedback}
\label{sec:sspatial_implementation}
Torso-tracking errors can displace the hands from their intended paths. Spatial feedback preserves each wrist target in the local odometry frame $O$ and re-expresses it using the measured torso pose at execution time $t$:
\begin{equation}
 \trans{B_t}{G_h^\star(t)}=
 (\trans{O}{B_t})^{-1}\trans{O}{B_o}\trans{B_o}{G_h^\star(t)}.
 \label{eq:sspatialtarget}
\end{equation}
This lets the arms compensate for body-tracking deviations while retaining the original spatial wrist targets.

Forward kinematics gives the measured wrist poses. We compute position and rotation errors against the re-expressed targets, together with velocity errors that include measured body and joint motion. An arm-Jacobian least-squares solve, with damping, posture regularization, and bounded updates, produces corrections to the controller's arm-joint targets. Only the 14 arm joints are corrected; the legs and waist retain the whole-body controller output. The policy remains fixed during this correction. Section~\ref{sec:spatial_ablation} evaluates its effect.

\subsection{Inference latency}
Online object perception and policy inference take 75.0\,ms and 251.8\,ms per update on average, respectively, during steady-state deployment. Perception includes mask updates and object-pose estimation, while policy inference includes RTC. We schedule replanning requests at a nominal interval of 0.4\,s.

\FloatBarrier
\section{Evaluation details and additional results}
\label{sec:sprotocol}
\subsection{Training data and controlled comparisons}
Each task uses a separate policy post-trained on 150 human demonstrations. Robot training data match the number $F$ of unique valid human observation frames under the common 30\,Hz convention. Episode counts may differ between sources.

All variants share the pretrained foundation, action interface, normalization, and optimization budget. Within each data source, policies with and without affordances are trained separately on identical frames, targets, and masks. They receive the same current pose and opening-state information. The controller and execution settings are shared, with spatial feedback enabled throughout the main-paper comparisons. The main comparison evaluates complete pipelines that differ in both data source and affordance input; within-source ablations isolate affordance conditioning.

\subsection{Affordance conditioning across data sources}
\label{sec:s_affordance_ablation}
Table~\ref{tab:s_policy_results} consolidates the main-paper TPS results under the shared execution setting. These scores also provide the feedback-on reference for the supplementary ablation below.

\begin{table}[t]
\centering
\caption{\textbf{Main-paper TPS with spatial feedback enabled.} $+$/$-$ denote with/without affordances. Ego$+$ is \egoloaf{}; Robot$-$ is the teleoperation baseline.}
\label{tab:s_policy_results}
\small
\setlength{\tabcolsep}{4pt}
\begin{tabular*}{\linewidth}{@{\extracolsep{\fill}}lrrrr@{}}
\toprule
Task & Ego$-$ & Ego$+$ & Robot$-$ & Robot$+$\\
\midrule
Tissue-to-Box & 64.2 & 81.0 & 81.4 & 90.0\\
Tissue-to-Bin & 55.8 & 70.0 & 75.2 & 80.0\\
Pick-Turn-Place & 50.2 & 75.4 & 74.8 & 100.0\\
Tissue-to-Cabinet & 42.5 & 65.8 & 66.0 & 75.0\\
\midrule
Four-task mean & 53.2 & 73.1 & 74.4 & 86.3\\
\bottomrule
\end{tabular*}
\end{table}

\subsection{Spatial-feedback ablation}
\label{sec:spatial_ablation}
\begin{figure}[t]
\centering
{\footnotesize
\begin{tabular}{@{}ll@{}}
\textcolor[HTML]{2878A6}{\rule{1em}{0.65em}}\enspace Full configuration &
\textcolor[HTML]{D9824C}{\rule{1em}{0.65em}}\enspace w/o affordances\\
\textcolor[HTML]{549D83}{\rule{1em}{0.65em}}\enspace w/o SF &
\textcolor[HTML]{A5A9AE}{\rule{1em}{0.65em}}\enspace w/o both
\end{tabular}\par}
\smallskip
\includegraphics[width=0.85\linewidth]{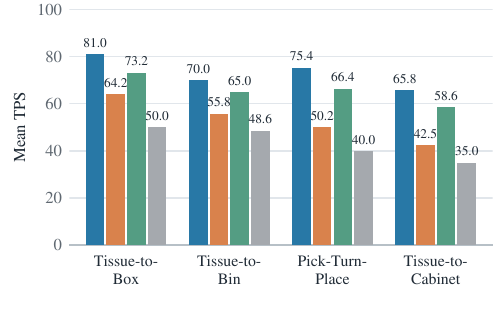}
\caption{\textbf{Human-data affordance/feedback ablation.}
Mean TPS (0--100, $\uparrow$) on all four tasks. SF denotes spatial feedback. Removing affordances uses a separately trained policy; toggling SF reuses its checkpoint.}
\label{fig:s_human_ablation}
\end{figure}

Figure~\ref{fig:s_human_ablation} evaluates spatial feedback on all four tasks, following the TPS rubric and evaluation protocol in Sec.~\ref{sec:experimental_setup} of the main paper. For each affordance setting, feedback-on and feedback-off runs reuse the same checkpoint with all other execution settings fixed. Disabling spatial feedback removes only the additional correction in Sec.~\ref{sec:sspatial_implementation}; ordinary closed-loop control remains enabled.

Spatial feedback improves TPS in both policy variants on all four tasks. Affordance conditioning remains beneficial when the additional correction is disabled, yielding gains of 23.2, 16.4, 26.4, and 23.6 points on Tissue-to-Box, Tissue-to-Bin, Pick-Turn-Place, and Tissue-to-Cabinet, respectively.

\subsection{Data-collection efficiency}
\label{sec:collectionprotocol}
\label{sec:scollection_summary}
We measure acquisition effort in person-hours~\citep{umi}, the working time summed across participants. This includes setup, demonstration collection, failed attempts, and resets, excluding unattended processing and post-collection annotation/review. Egocentric collection uses one person per task, while robot teleoperation uses two.

Table~\ref{tab:collection_results} reports collection duration, staffing, and person-hours for matched training-frame budgets. Person-hours equal duration in hours multiplied by staffing; the teleoperation-to-ego person-hour ratio gives each task's efficiency gain. The equal-task mean of these ratios is $4.03\times$.

\begin{table}[t]
\centering
\caption{\textbf{Measured acquisition effort.} Matched training-frame budgets. Gain is the teleoperation-to-ego person-hour ratio.}
\label{tab:collection_results}
\small
\setlength{\tabcolsep}{3pt}
\begin{tabular*}{\linewidth}{@{\extracolsep{\fill}}llrrrr@{}}
\toprule
Task & Source & People & Minutes & Person-h & Gain\\
\midrule
Box & Ego & 1 & 30 & 0.500 & \multirow{2}{*}{$2.33\times$}\\
 & Teleop. & 2 & 35 & 1.167 & \\
Bin & Ego & 1 & 37.5 & 0.625 & \multirow{2}{*}{$3.20\times$}\\
 & Teleop. & 2 & 60 & 2.000 & \\
Plush & Ego & 1 & 42.5 & 0.708 & \multirow{2}{*}{$2.59\times$}\\
 & Teleop. & 2 & 55 & 1.833 & \\
Cabinet & Ego & 1 & 62.5 & 1.042 & \multirow{2}{*}{$8.00\times$}\\
 & Teleop. & 2 & 250 & 8.333 & \\
\bottomrule
\end{tabular*}
\end{table}

Tissue-to-Cabinet has the largest gain ($8.00\times$). Returning the humanoid to its starting position after long-distance movement is time-consuming, contributing to the larger collection-efficiency gap on this task.

\FloatBarrier
\section{Dataset coverage and organization}
\label{sec:srelease}
The \egoloaf{} dataset contains 128,400 independently recorded demonstrations across 1,284 tasks, with 100 demonstrations per task and 1,000 hours of video after quality filtering. This section details the task taxonomy, representative task sequences, and the organization of observations and annotations.

\subsection{Task taxonomy and coverage}
\label{sec:s_task_taxonomy}
The task inventory organizes manipulation into six families and 30 finer-grained operation types. These cover object placement and organization, opening and closing, control actuation, alignment and insertion, material transfer, and tool use on surfaces. Table~\ref{tab:s_task_taxonomy} lists the operation types and their coverage under three body-motion settings. The complete inventory of 1,284 tasks is provided in \emph{task\_catalog.xlsx} in the supplementary material, with task IDs linking the catalogue to the examples below.

Each task specifies the objects, initial arrangement, action sequence, and completion criterion. The count of 1,284 refers to these concrete task configurations, including configurations that combine a shared manipulation goal with different body-motion requirements. For example, placing a cup on a coaster appears as an in-place operation, an operation requiring a turn toward a side workspace, and an operation requiring the cup to be carried to another workspace. Each configuration has its own independent demonstrations.

The three settings distinguish how the body participates in manipulation. \textbf{In-place manipulation} contains 600 tasks performed within reach at one standing workspace. \textbf{Posture-coordinated manipulation} contains 400 tasks requiring turning, bending, squatting, or combinations of these movements within the same work area. \textbf{Loco-manipulation} contains 284 tasks that connect spatially separated workspaces through walking, such as carrying an object to a destination or retrieving a tool before using it elsewhere. This organization covers both local hand--object interactions and their coordination with torso movement.

\begin{table}[tbp]
\centering
\caption{\textbf{Task taxonomy and coverage.} The inventory groups 1,284 task configurations into six manipulation families and 30 operation types. A: in-place manipulation; B: posture-coordinated manipulation; C: loco-manipulation. Counts denote task configurations.}
\label{tab:s_task_taxonomy}
\small
\setlength{\tabcolsep}{4pt}
\renewcommand{\arraystretch}{1.12}
\begin{tabular*}{\textwidth}{@{\extracolsep{\fill}}>{\raggedright\arraybackslash}p{2.9cm}>{\raggedright\arraybackslash}p{9.0cm}rrrr@{}}
\toprule
Manipulation family & Operation types & A & B & C & Total\\
\midrule
Placement and organization & Surface-to-surface transfer; loading/unloading open containers; sorting and arranging; stacking/nesting; tray loading and bimanual carrying & 100 & 100 & 100 & 300\\
Opening and closing & Hinged doors; drawers; lids; latches and clips; combined opening, object transfer, and closing & 100 & 100 & 80 & 280\\
Control actuation & Buttons; toggle/slide switches; levers; rotary controls; sequential control operation and state restoration & 100 & 40 & 0 & 140\\
Alignment and insertion & Slot placement; peg/hole insertion and removal; connector mating; guided sliding, fitting, and hanging; structural assembly/disassembly & 100 & 80 & 22 & 202\\
Material transfer & Liquid pouring; granular pouring and recovery; scooping; portioning; mixing followed by transfer & 100 & 40 & 41 & 181\\
Tool and surface operations & Wiping; brushing; scraping/squeegeeing; sweeping and collection; pressing, rolling, and stamping & 100 & 40 & 41 & 181\\
\midrule
Total & & 600 & 400 & 284 & 1,284\\
\bottomrule
\end{tabular*}
\end{table}

\subsection{Representative task sequences}
\label{sec:s_task_examples}
Table~\ref{tab:s_task_examples} gives task sequences and observable completion criteria for pouring, object alignment, low-height operation, transport, and tool use. Each demonstration retains the transitions between grasping, changing body posture or location, interacting with the target, and releasing the object.

\begin{table}[tbp]
\centering
\caption{\textbf{Representative task specifications.} IDs match the complete task catalogue; A/B/C follow Table~\ref{tab:s_task_taxonomy}. A401 and C001 are illustrated in Fig.~\ref{fig:s_dataset_sequences} and the accompanying videos.}
\label{tab:s_task_examples}
\small
\setlength{\tabcolsep}{4pt}
\renewcommand{\arraystretch}{1.12}
\begin{tabular*}{\textwidth}{@{\extracolsep{\fill}}l>{\raggedright\arraybackslash}p{2.7cm}>{\raggedright\arraybackslash}p{7.8cm}>{\raggedright\arraybackslash}p{4.4cm}@{}}
\toprule
ID & Task & Action sequence & Completion criterion\\
\midrule
A401 & Pour water into an empty cup & Pick up the jug, align it above the cup, pour water into the cup, and return the jug to the table. & Water is in the cup and the jug rests on the table.\\
A321 & Insert a peg into a hole & Pick up the peg, align it with the hole axis, and insert it to the stop. & The peg reaches the stop and stands in the hole.\\
B003 & Move a vase at low height & Squat, support the vase at its body and base, transfer it to the mat, and stand up. & The vase remains upright on the mat.\\
B223 & Release a clamp at low height & Squat, rotate the screw-clamp knob to loosen the clamp, and remove the two cardboard pieces. & The clamp releases both pieces, which are removed intact.\\
C001 & Carry a cup to a coaster & Grasp the cup handle with one hand, walk to the destination, align the cup with the coaster, and place and release it. & The cup is upright with its base within the coaster.\\
C244 & Retrieve a cloth and wipe a tray & Take the cloth from the tool area, walk to the tray, wipe up the water, and put the cloth in the collection basket. & No visible pooled water remains; the cloth is in the basket.\\
\bottomrule
\end{tabular*}
\end{table}

\begin{figure}[tbp]
\centering
\includegraphics[width=\textwidth]{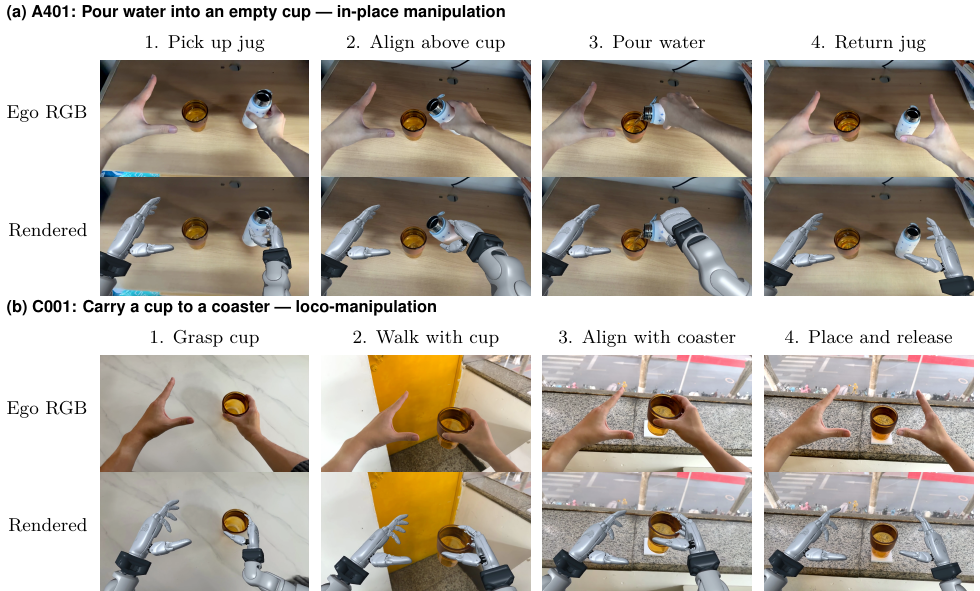}
\caption{\textbf{Temporal examples from the dataset.} Four successive stages of A401 and C001, with paired original and rendered observations at each selected time. The full sequences are provided in the accompanying supplementary videos.}
\label{fig:s_dataset_sequences}
\end{figure}

Figure~\ref{fig:s_dataset_sequences} follows two tasks through four stages, pairing the original egocentric view with the rendered robot view at each selected time. In \textbf{A401}, the demonstrator picks up a water jug, aligns it above an empty cup, pours water into the cup, and returns the jug to the table, all at one workspace. In \textbf{C001}, the demonstrator grasps a cup by its handle with one hand, carries it to another workspace, aligns it with a coaster, and places it down. The first example emphasizes local hand--object interaction, while the second connects grasping and placement through walking. Both use the same torso-and-hand annotation interface. Complete videos of A401 and C001 accompany the supplementary material.

\begin{table}[tbp]
\centering
\caption{\textbf{Dataset contents and learning roles.} Per-frame validity masks identify unavailable or unreliable annotation components.}
\label{tab:datasetfields}
\small
\setlength{\tabcolsep}{4pt}
\begin{tabular}{@{}>{\raggedright\arraybackslash}p{3.1cm}>{\raggedright\arraybackslash}p{6.6cm}>{\raggedright\arraybackslash}p{6.1cm}@{}}
\toprule
Field & Contents & Role in learning\\
\midrule
Source recording & RGB, timestamps, intrinsics, camera trajectory & Source observations and reconstruction\\
Task instruction & Language description of the complete demonstration & Task conditioning\\
Torso and hand motion & Virtual-torso and bilateral TCP trajectories & Current pose state and future action targets\\
Hand opening & Binary left/right opening states and validity & Current state and future opening targets\\
Object geometry & Object poses, hand--object affordances, validity & Current geometric conditioning\\
Rendered observations & G1/O6 appearance composited into source frames & Policy observations with robot appearance\\
\bottomrule
\end{tabular}
\end{table}

\subsection{Episode contents and quality control}
\label{sec:s_dataset_contents}
Each episode pairs its task instruction with the original egocentric recording, rendered robot observations, and geometric annotations. Table~\ref{tab:datasetfields} summarizes these fields and their learning roles. Language describes the complete demonstration; camera, torso, hand, and object annotations are aligned with individual observations. Original and rendered images share timestamps and geometric annotations, allowing the same action targets to be paired with either appearance.

Torso and bilateral TCP trajectories provide current pose states and future action targets, while hand-opening labels specify the corresponding opening states. Object poses and hand--object affordances provide geometric conditioning at the current observation. The continuous trajectories support different target horizons and temporal sampling without repeating reconstruction. Sections~\ref{sec:sacquisition} and~\ref{sec:straining} describe the conversion and training procedures.

We check demonstrations for recording completeness, camera-trajectory quality, arm and hand rendering quality, and task completion, discarding demonstrations that fail these checks. Per-frame masks preserve the validity of individual annotation components so that training can use the available supervision. Task-specific training and evaluation settings are given in Sec.~\ref{sec:sprotocol}.

\end{document}